\documentclass[10pt,twocolumn,letterpaper]{article}
\usepackage{amsmath,amssymb} 
\usepackage{subfigure}
\usepackage[ruled,vlined]{algorithm2e}
\usepackage{wacv}
\usepackage{times}
\usepackage{epsfig}
\usepackage{graphicx}
\usepackage{amsmath}
\usepackage{amssymb}
\usepackage{multirow}
\usepackage{soul}

\newcommand{\IGNORE}[1]{}

\def\argmin{\mathop{\rm arg\,min}\limits}

\wacvfinalcopy 

\def\wacvPaperID{233} 

\ifwacvfinal\fi
\begin{document}

\title{Real-time Online Action Detection Forests using Spatio-temporal Contexts}

\author{Seungryul Baek \\
Imperial College London\\
{\tt\small s.baek15@imperial.ac.uk}
\and
Kwang In Kim \\
University of Bath\\
{\tt\small k.kim@bath.ac.uk}
\and
Tae-Kyun Kim \\
Imperial College London\\
{\tt\small tk.kim@imperial.ac.uk}
}

\maketitle
\ifwacvfinal\thispagestyle{empty}\fi

\begin{abstract}
Online action detection (OAD) is challenging since 1) robust yet computationally expensive features cannot be straightforwardly used due to the real-time processing requirements and 2) the localization and classification of actions have to be performed even before they are fully observed. We propose a new random forest (RF)-based online action detection framework that addresses these challenges. Our algorithm uses computationally efficient skeletal joint features. High accuracy is achieved by using robust convolutional neural network (CNN)-based features which are extracted from the raw RGBD images, plus the temporal relationships between the current frame of interest, and the past and \emph{futures} frames. While these high-quality features are not available in real-time testing scenario, we demonstrate that they can be effectively exploited in training RF classifiers: We use these \emph{spatio-temporal contexts} to craft RF's new split functions  improving RFs' leaf node statistics. Experiments with challenging MSRAction3D, G3D, and OAD datasets demonstrate that our algorithm significantly improves the accuracy over the state-of-the-art online action detection algorithms while achieving the real-time efficiency of existing skeleton-based RF classifiers. 
\end{abstract}

\section{Introduction}
\label{sec:intro}


\IGNORE{
\begin{itemize}
\item Discussion on our main argument from the application scenario first
\end{itemize}}
\IGNORE{
\begin{itemize}
\item Use expensive but more powerful features in training.
\item look ahead in time, again in training.
\end{itemize}
\item Framework: how we use expensive features + temporal context (look ahead in time) in pair-wise and higher-order potential context.
\item Define `context'. 
\item Low dimensional features $\to$ real-time requirement (cheap features); Look ahead in time.
\item Equations (2-5): make it formal.}
\IGNORE{
\item Why unified? We perform action detection and classification using a single random forest.
}

\IGNORE{Please define online action detection. BTW, is it common to call it action detection (instead of recognition)?}
\IGNORE{This is a good intro. sentence. Can we use (move) the last part `even before the action is fully completed' in the discusssion of challenges?}

Online action detection (OAD) refers to the task of simultaneously localizing actions and identifying their classes from \emph{streamed} input sequences~\cite{OAD,SoftRegression,Sharaf15,online_eccv_2016,movingpose_2013}. OAD facilitates a wide range of real-time applications including visual surveillance, pervasive health-care, and human-computer interaction. 
These new opportunities, however come with significant challenges not encountered by the classical \emph{offline} action recognition~\cite{online_cvpr_2016,offline2,Gkioxari_2015_cvpr,biD1,Xia_cvor_2012,Jiang_tpami_2014,movingpose_2013,lie_group,NIPS2014_twostream,offline1,CVPR15_heterogeneous} approaches: Realistic OAD applications require real-time processing that prevents the use of computationally expensive features.  \IGNORE{Furthermore, the intended action has to be localized and classified immediately after or even before the action is completed.} Significant progress in this direction has been made with the the recent emergence of consumer-level depth sensors: For instance, the Microsoft Kinect sensor enables real-time estimation of skeleton joints which provides highly descriptive and compact (\ie, $10$ to $20$ key-point coordinates of a human body) representations of complex human motions, and thereby facilitate subsequent real-time recognition.
However, due to occlusions \cite{skkdiff2} and variations of appearance including clothing, body shape, and scale~\cite{skkdiff1}, estimated skeleton joints are often noisy~\cite{Omar_cvpr_2013,hopc,g3d},\IGNORE{estimation is often unreliable } \IGNORE{this part doesn't seem to be informative, what do we mean by `freely available' and `additional shape information'? We have to `characterize' RGBD features from our perspective. Also, is it about RGBD features or RGBD features + CNNs?} leading to unreliable action recognition.\IGNORE{erroneous the entire skeleton-based OAD framework.} \IGNORE{Please note that the MS Kinect sensor is an RGBD sensor, all Kinect-based algorithms including skeleton-based algorithms are RGBD sensor-based.}

\IGNORE{Not sure if `fine-grained' is compatible with `dealing with large intra-class variations'.}

On the other hand, the state-of-the-art convolutional neural network (CNN)-based features~\cite{Alex_nips_2012,devil_bmvc} extracted from RGB and depth images have demonstrated greater robustness in action recognition~\cite{novel3d,nktm,CVPR2016_twostream,NIPS2014_twostream}. In particular, they are known to be effective in capturing the fine-grained attributes of human actions~\cite{deepattr1} (as well as other scene objects \cite{deepattr2}) providing robustness to appearance variations \cite{skkdiff1,movingpose_2013} often encountered in human action recognition. However, incorporating sophisticated CNN features in OAD frameworks is challenging due to the real-time processing requirement.

Another main challenge of OAD is that the intended action has to be localized and classified immediately after or even before the action is completed. This precludes exploiting temporal context of the current action spanning the previous and the unobserved future frames~\cite{biD1,biD2}.

In this paper, we present a new framework that effectively incorporates the expensive CNN-based features as well as the temporal context present in the past and future frames. The key idea is to exploit these additional \emph{spatio-temporal contexts} only in the \emph{training} phase of the action detection system. Once trained, our system uses the computationally efficient skeleton features enabling online, real-time testing. We instantiate this idea in a novel random forests (RF) algorithm where the refined RF parameters (\ie split functions and leaf node statistics) are learned with the aid of contexts. The use of RFs are motivated by the facts that 1) Each tree of RFs is computationally extremely efficient and  independent (easily parallelizable), which make RFs suitable for real-time applications; 2) RF is highly flexible, accommodating multiple feature modalities (\eg skeletons, CNN-based features, and time indices) and the heterogeneous training criteria (regression and classification) allowing the incorporation of both action localization (regression) and classification into a single unified RF framework.

Experimental comparison with eight state-of-the-art algorithms on three challenging datasets demonstrate that our new algorithm achieves the state-of-the-art performance while retaining the low computational complexity suitable for OAD.

Our contributions to the literature are summarized as:
\begin{itemize}
\item We propose to use RGBD-based CNN features as spatial contexts and temporal location features as temporal contexts to compensate for the noisy skeleton estimation and lack of sufficient temporal information in the OAD framework.
\item We instantiate spatio-temporal contexts in RFs using the objective similar to the conditional random field. Instead of using the original objective, we divide them into independent multiple objectives and randomly mix them up sequentially in the RFs.\IGNORE{Is it a significant contribution?}
\item We obtain real-time action detection forests by mapping all contextual information by improving split functions of RFs and resultant leaf node statistics at the training stage, which makes the contexts not used at the testing stage for the real-time efficiency.
\end{itemize}

\section{Related works}
This section reviews related works, categorized into two: 1) skeleton-based OAD; 2) random forests that exploits auxiliary information. Our RF framework can be seen as a particular case of the second category.  
\label{s:relatedwork}
\paragraph{Skeleton-based online action detection.}
\IGNORE{The point of this sentence is?: }

Skeleton-based OAD frameworks \cite{G3Dtest,movingpose_2013,ap_tr_2012,realtime_wacv_2015,Sharaf15,online_eccv_2016} has been developed so far. In \cite{ap_tr_2012}, short sequences (\ie $35$-frame interval sequences) were defined as {\it action points} based on skeleton joints and they showed that this basic unit and RF-based framework is promising for the online method, due to good performance and efficiency. In \cite{G3Dtest}, {\it action points} were annotated for the G3D dataset proposed in \cite{g3d}. A similar notion of action points was utilized in \cite{realtime_wacv_2015} but with different-scale time intervals. The moving pose descriptor \cite{movingpose_2013} achieved a good performance by capturing both the skeleton joints at one frame and the speed/acceleration of joints around the frame in a short sequence. In \cite{Sharaf15}, the combination of the moving pose descriptor using the angles and the codebook model showed the promising results in real-time online action recognition. In \cite{online_eccv_2016}, Li \etal introduced more realistic OAD datasets and baseline methods. They proposed to use the recurrent neural network (RNN)-based joint classification and regression framework using raw skeleton inputs and have shown the state-of-the-art results with relatively low time complexity. They reported that skeleton input is much attractive than RGBD-based \cite{OAD} for OAD framework due to computational efficiency. 

As noted in \cite{OAD,online_eccv_2016}, OAD is a challenging problem and even though skeleton-based approaches \cite{G3Dtest,movingpose_2013,ap_tr_2012,realtime_wacv_2015,Sharaf15,online_eccv_2016} are promising, there is a room for improvement by combining multiple feature modalities and considering more temporal relationships. Dissimilar to the previous approaches \cite{ap_tr_2012,realtime_wacv_2015,Sharaf15,online_eccv_2016} that depend on a limited feature space, we try to explore multiple spatio-temporal feature spaces but only at the training stage, to improve the classification accuracy while keeping the real-time efficiency of the testing stage.

\IGNORE{Overall, the related work section about putting our work and existing work in contexts. Why and how the methods discussed in this paragraph are relevant?}

\paragraph{Training forests with data relationships.}
\IGNORE{Please help readers to be prepared for what comes next. So far we haven't every talked about `dependencies'. Why does it come now? Is this term somehow used in place of `contexts'?:}
\IGNORE{Just from curiosity, I was wondering why we have to classify existing RFs with `local' and `global' contexts when there's only one paper that uses global contexts?}
There have been several works \cite{hrf_iccv2013,pairwise_rf1,pforest,geof,entangle,stf} that use the auxiliary information when training RFs other than the mapping between feature space to the label space. Some use pairwise relationships between data samples \cite{hrf_iccv2013,pairwise_rf1,stf,entangle} while others \cite{pforest} use global information when training RFs. Also, there have been methods combining conditional random fields and RFs to use both pairwise and global information in a forest \cite{geof}. In \cite{hrf_iccv2013}, pairwise links between synthetic and natural dataset were used to penalize separating those data in each split for the purpose of transfer learning. In \cite{pairwise_rf1}, to deal with sparse and noisy information for the clustering problem, pairwise links were proposed between samples as hard constraints (\ie must-link, cannot-link) and forests stopped growing when split functions violate the constraints. In \cite{pforest}, a face recognition problem was dealt with by considering a head pose as a global variable at the training stage. Since continuous head poses were hard to treat in their framework, they discretized the head pose space into several discrete labels and incorporated it as another classification task.\IGNORE{In this sentence, does `long range dependencies' come as a part of `the works using local contexts'? I wonder if `long range' and `local contexts' sound contradictory.} In \cite{stf,entangle}, to incorporate both label co-occurrency and consistency that is important for the segmentation task, new feature encoding method was designed for RF inputs. \cite{entangle} is differentiated from \cite{stf} by performing it in an unified forest. \cite{geof} is generalized approach for the \cite{entangle,stf} and they proposed new feature encoding method and a field-inspired split objective for RFs to incorporate the spatial consistency. However, \cite{geof} reported that they did not fully use pairwise and global higher-order data relationships between samples since it does not always improve the classification accuracy. As a result, their spatial consistency is come from feature encoding methods rather than the split function. 

We train our RFs by using conditional random field-like objective for structurally predicting RFs’ split parameters. Compared to previous algorithms, we fully consider pairwise and higher-order potentials and use multi-modal feature spaces for describing data relationships. Also, to secure the real-time testing time for our application, we only use the multi-modal feature space at the training stage only.

\section{Exploiting contexts in skeleton-based OAD}
\label{s:exploitingcontexts}
\IGNORE{Can we just say `(Exploiting) contexts in online detection' to stress the role of `context'? : )}
\label{sec4}
\IGNORE{I would prefer placing this section (perhaps section 4.2) before section 3. While I was reading section 3 I had to constantly ask why? This could be answered straightforwardly by properly defining `context': What it means, why it helps. How it is formally represented.}
This section defines the spatio-temporal contexts and explains how they can complement the skeletal features. Section~\ref{s:oadforestswithcontexts} details instantiating the resulting context-guided action recognition framework with random forests.

\IGNORE{In this section, we explain the feature descriptor and context information that we use for the action recognition.}

\IGNORE{We could combine the following two paragraphs? Do we need separate headings \textbf{Local Descriptor} and \textbf{Optimization}?}

\subsection{Skeleton-based feature space}
Skeleton joints are widely used in the real-time action recognition as they provide efficiency in feature extraction~\cite{ap_tr_2012,realtime_wacv_2015,Sharaf15,movingpose_2013,online_eccv_2016}. We use a combination of skeletal joint locations and their time derivatives proposed by Zanfir~\etal~\cite{movingpose_2013}: A $n\times 3\times 3$-dimensional feature vector $\mathbf{x}\in \mathcal{X}$ is formed by concatenating the $n$ joint positions $\mathbf{p}(t)$ consisting of $x$, $y$ and depth coordinate values, and their first and second order time derivatives, $\dot{\mathbf{p}}(t)$ and $\ddot{\mathbf{p}}(t)$, respectively: $\mathbf{x}=[\mathbf{p}^\top,\dot{\mathbf{p}}(t)^\top,\ddot{\mathbf{p}}^\top]^\top$.

\subsection{Spatio-temporal contexts}
\IGNORE{My understanding is that roughly the argument is 1) In OAD, information in future frames is not available at the testing stage, and due to the real-time processing (feature extraction and classification) requirement, computationally expensive features cannot be used. 2) We use this \emph{lost} information as contexts that can be exploited in training RFs (Sec.~\ref{s:trainingforests}). }
In OAD, information on future frames is not available at the testing stage, and due to the real-time processing requirement, computationally expensive features cannot be used. We exploit these powerful but unavailable auxiliary features to guide the \emph{training} process, by defining them as a context feature space $\mathcal{Z}=
\{\mathcal{Z}_S,\mathcal{Z}_T\}$ that is composed of spatial and temporal feature spaces: $\mathcal{Z}_S$ and $\mathcal{Z}_T$. The context feature space contains a context feature vector $\mathbf{z}(I)=[\mathbf{z}_S(I)^{\top},\mathbf{z}_T(I)^{\top}]^{\top}\in\mathcal{Z}$ for each frame $I$, where $\mathbf{z}_T(I)\in\mathcal{Z}_T$ and $\mathbf{z}_S(I)\in\mathcal{Z}_S$. Specific instances of these spatio-temporal contexts (as detailed in the remainder of this section) are motivated by their success in the corresponding offline applications \cite{NIPS2014_twostream,CVPR2016_twostream} while other contexts are possible. One of our main contributions is to demonstrate that these offline application-motivated features can be effectively used in the training stage (See~Sec.~\ref{s:oadforestswithcontexts}).


\paragraph{Spatial contexts.}\IGNORE{Redundant from here} Our spatial context is defined as the expensive yet powerful CNN feature descriptor applied on the holistic scene where human actions are performed. For both depth and RGB maps, we extract deep architecture-based $4,096$-dimensional feature responses for each frame $I$. Then, we concatenated their $fc7$ responses and define the $8,192$-dimensional spatial context vector $\mathbf{z}_S(I)\in \mathcal{Z}_S$. We used different architectures for RGB and depth map, respectively. For the RGB map, we used the {\it VGG-s} \cite{devil_bmvc} architecture, pre-trained on ImageNet while for the depth map, we used architecture proposed in \cite{novel3d}, pre-trained on $180$-different viewed depth maps.  
\paragraph{Temporal contexts.} 
\IGNORE{I would like to an even more intuitive argument. We mentioned previously that we are detecting and classifying actions in streamed sequences. Could you describe in plain language (without any math) what information we aim to extract (\ie what we mean by temporal coherency)? Is it about how a frame (or features observed at a specific time instance) can be put in the context of the entire action sequence (still very vague though)?:}
Our temporal context is defined as the relative temporal location of a frame of interest in the entire action sequence that generates the frame. This context type is motivated by its ability to disambiguate similar-looking features: Some (similar) features are inherently ambiguous in the sense that multiple action classes can exhibit them in the corresponding action sequences. However, these features are often observed in different relative times in the entire action sequences and therefore, putting them in this context can help reveal the underling action classes. Since this temporal context is not available in testing, it cannot be used in disambiguating similar features. Instead, exploiting it in training enables putting emphasis on refining decisions on these ambiguous features in $\mathcal{X}$. Given an action sequence $\mathcal{A}$, the temporal context $\mathbf{z}_T(I)\in \mathcal{Z}_T$ of a frame $I\in\mathcal{A}$ is defined as follows:
\begin{eqnarray}
\mathbf{z}_T(I)=\frac{\mbox{loc}_{\mathcal{A}}(I)}{\mbox{length}(\mathcal{A})}
\end{eqnarray}
where $\mbox{loc}_\mathcal{A}(\cdot)$, $\mbox{length}(\cdot)$ denote the frame index in the action sequence $\mathcal{A}$ and the length of the action sequence $\mathcal{A}$. 
\IGNORE{Why RFs here?}

\section{Online action detection forests with contexts}
\label{s:oadforestswithcontexts}
Our random forests adopt two types of features: the skeleton features $\mathbf{x}$ and spatio-temporal features $\mathbf{z}=(\mathbf{z}_S,\mathbf{z}_T)$. Both feature types are used in training (\ie constructing the split functions of the RFs) to exploit the rich contextual information. However, in testing, only skeleton features are used as the spatio-temporal contexts are not available.




\paragraph{Random forests.} Random Forests (RFs) are ensembles of binary decision trees that are constructed based on bootstrapped data. Each tree consists of two types of nodes: \emph {split nodes} and \emph{leaf nodes}. For an input data point $I$ and the corresponding feature representation $\mathbf{h}(I)$, a split node conducts a simple binary test (\eg by comparing the $\gamma$-th feature value $\mathbf{h}_{\gamma}(I)$ of $\mathbf{h}(I)$ with a certain threshold $t$). Each out-branch of the split node then represents the outcome of the test, and $I$ is accordingly sent to either the left or right child (\eg left if $\mathbf{h}_{\gamma}(I)\le t$, right otherwise). During training, RFs iteratively partition the training set based on the split nodes. Therefore, the leaf node arrived by traversing $I$'s path from the root represents an empirical estimate of the class-conditional distribution (or posterior distribution in general, \eg for regression) that matches to the tested features of $I$. The final classification outputs of RFs are calculated by taking majoring voting from the modes of individual tree distributions. Regression outputs are similarly calculated.

Training RFs involves growing the trees by deciding each node's decision behavior as prescribed by a \emph{split function} $\Psi$: The split function of each node is defined as $\Psi(\mathbf{h}_{\gamma}(I),t) := \text{sgn}(\mathbf{h}_{\gamma}(I)-t)$.\footnote{For computational efficiency, we only consider simple single-feature comparisons while more complex split functions are also possible.}
Once the input training (sub-)set $\mathcal{D}$ is 
arrived at each node, a set of candidate split functions $\{\Psi^c\}$ is randomly generated and among them, the one that maximizes a split objective function $\mathcal{O}$ is selected as a final split function $\Psi^*$:
\begin{eqnarray}
\Psi^{*}=\argmin_{\Psi\in\{\Psi^c\}} \mathcal{O}(\Psi)
\end{eqnarray}
which divides $\mathcal{D}$ into two disjoint subsets $\mathcal{D}_L$ and $\mathcal{D}_R$. The objective function $\mathcal{O}$ can measure the average reduction of the class entropy (\ie training error or information gain) or it can capture certain higher-order smoothness in the splitting process (Eqs.~\ref{e:efirst}-\ref{e:ethird}). 

\paragraph{Online action detection with random forests.} 
A major advantage of using the spatio-temporal contexts over skeleton features comes from their higher discriminative capacity. The underlying rationale of using these contexts only in training RFs (as they are not available in testing) is that they can steer the construction of split functions to put more emphasis on making fine-grained decisions on \emph{ambiguous} patterns (the data points similar to each other in $\mathcal{X}$ but with different class labels). Our preliminary experiments have revealed that spatio-temporal contexts can be best exploited when they are directly used to refine the geometry of the underlying data space: Even if $\mathbf{x}(I_j)$ and $\mathbf{x}(I_k)$ are similar, they should belong to different tree branches when $\mathbf{z}(I_j)$ and $\mathbf{z}(I_k)$ are different.

To incorporate the spatio-temporal contexts in such a way into RFs, we introduce a new split objective $\mathcal{O}$ which is similar to conditional random fields (CRFs) with higher-order potentials~\cite{hiercrf1,hiercrf2,hiercrf3}:
\begin{eqnarray}
\label{e:crfpotential}
\mathcal{O}(\Psi;\mathcal{Z}) = \underbrace{\mathcal{O}_{u}(\Psi)}_{unary}+\lambda_1\underbrace{\mathcal{O}_{p}(\Psi;\mathcal{Z})}_{pairwise}+\lambda_2\underbrace{\mathcal{O}_{h}(\Psi;\mathcal{Z})}_{higher-order},
\end{eqnarray}
where $\lambda_1$ and $\lambda_2$ are a hyper-parameters that weights contributions of the pairwise potential ($\mathcal{O}_{p}$) and the higher-order potential $\mathcal{O}_{h}$, and
\begin{align}
\label{e:efirst}
\mathcal{O}_{u}(\Psi) &= -\sum_{v\in\{L,R\}} \sum_{y\in\mathcal{Y}} n(y,\mathcal{D}_v) \mbox{log}\frac{n(y,\mathcal{D}_v)}{|\mathcal{D}_v|},\\
\label{e:esecond}
\mathcal{O}_{p}(\Psi;\mathcal{Z}) &= \sum_{v\in\{L,R\}}\sum_{I_j,I_k\in\mathcal{D}_v} ||\mathbf{z}(I_j)-\mathbf{z}(I_k)||_2,\\
\label{e:ethird}
\mathcal{O}_{h}(\Psi;\mathcal{Z}) &= \sum_{v\in\{L,R\}}\sum_{I_j\in \mathcal{D}_v} ||\mathbf{z}(I_j)-\frac{1}{|\mathcal{D}_v|}\sum_{I_l\in\mathcal{D}_v} \mathbf{z}(I_l)||_2,
\end{align}
where $n(y,\mathcal{D}_v)$ denotes the number of samples having class label $y$ in $\mathcal{D}_v$ and $\mathcal{D}_L$, $\mathcal{D}_R$ are candidate left, right data splits decided by the split function $\Psi$.\IGNORE{The above equation has to be mathematically defined.} Minimizing Eq.~\ref{e:crfpotential} offers a structured prediction for $\Psi$ and the CRFs with higher-order potentials \cite{hiercrf1,hiercrf2,hiercrf3} are well-studied and known to offer a more accurate structured prediction than pairwise CRFs \cite{paircrf}, by capturing the global consistency.

Typically in CRFs, the combined heterogeneous potentials (as in Eq.~\ref{e:crfpotential}) are jointly optimized~\cite{hiercrf1,hiercrf2,hiercrf3}. However, this requires explicitly tuning the hyper-parameters $\lambda_1$ and $\lambda_2$. Recently, \cite{hough,hrf_iccv2013,pforest} demonstrated that by randomly sampling a split objective at each node's splitting process, multiple split objectives can be straightforwardly incorporated in RFs, even without having to introduce the weighting hyper-parameters (\ie $\lambda_1$, $\lambda_2$). We optimize $\mathcal{O}$ (Eq.~\ref{e:crfpotential}) adopting this sampling approach.

\IGNORE{We use the context feature space $\mathcal{Z}=\{\mathcal{Z}_S,\mathcal{Z}_T\}$ in Eq. 5 and Eq. 6 to incorporate them in the RFs. In general, three heterogeneous potentials in Eq. 3 can be jointly optimized as in conditional random fields \cite{hiercrf1,hiercrf2,hiercrf3}; however it requires explicit tuning of hyper-parameters $\lambda_1$ and $\lambda_2$. For RFs, it is well known that heterogeneous multiple split objectives can be mixed up by randomly sampling one of split objectives and deciding each node's split function $\Psi(\mathbf{x}_{\gamma}(\cdot),t)$ according to the sampled objective, when growing trees \cite{hough,hrf_iccv2013,pforest}. This bypasses the hyper-parameter tuning issue and leads to practically good results. This is one of good characteristics of RFs that makes RFs perform multiple types of tasks in one forest. }

As a result, we constitute total $5$ types of split objectives as $\mathcal{O}^{new}(\Psi;\mathcal{Z})$ to optimize RFs. Among $\mathcal{O}^{new}(\Psi;\mathcal{Z})=\{\mathcal{O}_{u}(\Psi),\mathcal{O}_{p}(\Psi;\mathcal{Z}_S),\mathcal{O}_{p}(\Psi,\mathcal{Z}_T),\mathcal{O}_{h}(\Psi,\mathcal{Z}_S),\mathcal{O}_{h}(\Psi,\mathcal{Z}_T)\}$, one of split objectives are randomly sampled at each node splitting, and split function $\Psi(\mathbf{x}_{\gamma}(\cdot),t)$ that minimizes sampled split objective is selected from split function candidates $\{\Psi^c\}$. The overall training process is summarized in Algorithm 1 and we detail the optimization process in each split objective in both Figure 1 and remainder of this section.

\begin{algorithm}[t]
\SetAlgoLined
$\mathbf{Input}$: $\mathcal{Z}=\{\mathcal{Z}_S,\mathcal{Z}_T\}$ and $\mathcal{X}$ for $\mathcal{D}$ arrived at each split node.\\
$\mathbf{Output}$: Split function $\Psi^*$, Data splits $\mathcal{D}_L$ and $\mathcal{D}_R$ at each split node.\\
   Generate a reference split $\Psi^0$ whose left child has all samples $\mathcal{D}_L=\mathcal{D}$ and right child has no sample $\mathcal{D}_R=\phi$.\\
   Randomly generate split function candidates $\{\Psi^c\}$ and corresponding data splits $\mathcal{D}_L$ and $\mathcal{D}_R$.\\
   Randomly select one of split objectives from $\mathcal{O}^{new}(\Psi;\mathcal{Z})$ defined in Sec 4.\\
   Calculate the score $\mathcal{O}(\Psi)$ for $\{\Psi^c\}$ and $\Psi^0$ according to the selected split objective (Eqs. 4-6).\\
   Find the Best Split function $\Psi^*$ by Eq. 2.\\
   Calculate the Information Gain by $\mathcal{O}(\Psi^*)-\mathcal{O}(\Psi^0)$.\\
   \If {No Information Gain on $\Psi^*$}
   {
       Make Leaf (\ie Stop Growing).\\
   }
   \Else
   {
      Split $\mathcal{D}$ into $\mathcal{D}_L$ and $\mathcal{D}_R$ using $\Psi^{*}$.\\          
   }
\caption{\label{alg:weight}Training each split node with three split objectives $\mathcal{O}=\{\mathcal{O}_{unary},\mathcal{O}_{pair},\mathcal{O}_{higher}\}$.}
\end{algorithm}

\paragraph{Unary objective: mapping skeletons to action class.}
The unary objective in Eq. 4 is equal to the objective of the standard RF method and measures the uncertainty of class label distributions in two splitted datasets $\mathcal{D}_L$ and $\mathcal{D}_R$. The formula in Eq. 4 denotes the sum of entropy measure in left and right child nodes. Among split function candidates $\{\Psi^c\}$, the one that minimizes the entropy measure (\ie Eq. 4) is selected as $\Psi^*$. This enforces to find a split function that has separated class label distributions in the left and the right child nodes as in Figure 1b. 

\paragraph{Pairwise objetive: between frame context consistency.}
\IGNORE{I guess here an important question is which features we are using: connected to the formal defintion of `context'.}
The pairwise objective in Eq. 5 measures the pairwise smoothness of data samples in a context feature space $\mathcal{Z}$. We borrowed the normalized cut (NCut) criterion, which transforms features into a Laplacian space where pairwise relationships can be well represented as in Figure 1c. 

The weights of edges $\mathcal{E}$ capture the pairwise smoothness between two trainig samples $I_j$ and $I_k$ in the contextual feature space $\mathcal{Z}$. The weights of edges $\mathcal{E}$ are defined as follows:
\begin{eqnarray}
\mathcal{E}(I_j,I_k;\mathcal{Z})=\exp\bigg(\frac{-||\mathbf{z}(I_j)-\mathbf{z}(I_k)||_2}{\sigma}\bigg)
\end{eqnarray}
where $\sigma$ is a parameter and we set its value as the mean of $||\mathbf{z}(I_j)-\mathbf{z}(I_k)||_2$ for $\forall I_j,\forall I_k \in \mathcal{D}$.

\begin{figure*}[t!]
 \centering
   \includegraphics[width=0.95\textwidth]{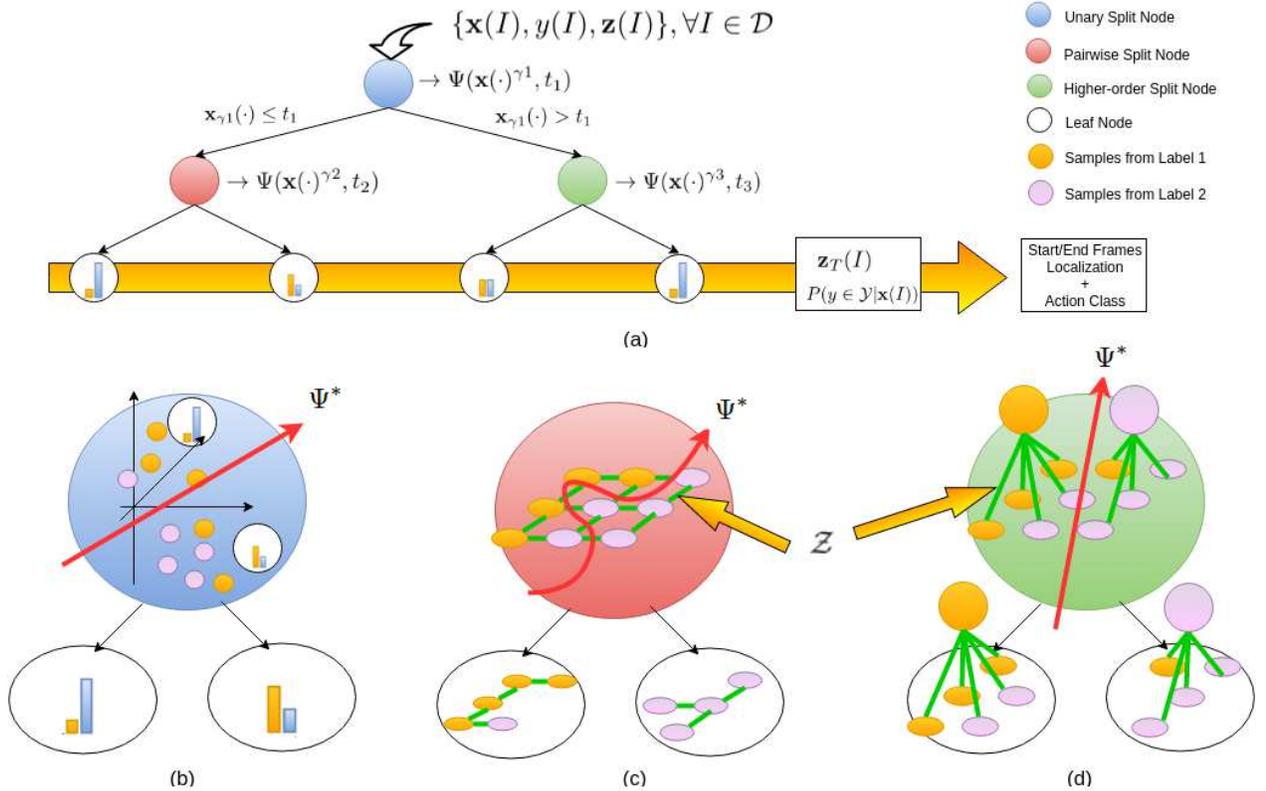}
   \label{fig:subfig3}
   \caption{Training stage of our algorithm for each tree. Trained forests are ensembles of trees. At the training stage, (a) split functions for every split nodes $\Psi(\mathbf{x}_{\gamma}(\cdot),t)$ are decided for each tree to learn a mapping from the feature space $\mathcal{X}$ to the label space $\mathcal{Y}$ with the aid of context feature space $\mathcal{Z}$. Note that at the testing stage, only the feature extractor $\mathbf{x}_{\gamma}(\cdot)$ is required, since learned split functions depend only on the feature space $\mathcal{X}$. Leaf nodes are terminating nodes and store class label distributions $P(y\in\mathcal{Y}|\mathbf{x}(I))$ as well as mean of location feature $\mathbf{z}_T$ (White circle). They are used for inferring action classes and localizing start/end frames. Our training objective is composed of unary, pairwise and higher-order potentials and this is optimized by three types of split objectives in the RFs. In (b), (c) and (d), we illustrate data splits using each split objective. Green lines in (c) and (d) denote context relationships based on context space $\mathcal{Z}$. (b) The unary objective is same as the standard entropy-based classification objective $O_{unary}$ (Blue circle) that enforces separate child nodes' class distributions. (c) The second is the proposed NCut-based split objective $O_{pair}$ that considers the pairwise relationships between samples based on context space $\mathcal{Z}$ (Red circle). (d) The third is the higher-order split objective $O_{higher}$ that considers relationships among more than 2 samples, by regarding mean of two groups as a global variable. In this objective, the consistency between mean of two groups and individual sample is considered based on context space $\mathcal{Z}$ (Green circle).}

\end{figure*}

The splits of RFs are inherently similar to the graph cuting \cite{ncut_2000,mincut} in a sense that splitted groups have no interaction ($\mathcal{D}_L\cap\mathcal{D}_R=\phi$) and the union of two groups is the entire set ($\mathcal{D}_L\cup\mathcal{D}_R=\mathcal{D}$). However, RFs' split objectives have not been considered for optimizing sum of pairwise similarities between samples as in Eq. 5, while objectives of graph cutting algorithms explicitly use pairwise similarities between samples. The direct optimization of Eq. 5 corresponds to the minimum cut criterion \cite{mincut} which has a bias to favor small sets of isolated nodes in the graph \cite{ncut_2000}. Instead, we follow the NCut criterion proposed in \cite{ncut_2000} that solved the issue. 

Splitting data into two groups $\mathcal{V}_A$ and $\mathcal{V}_B$ by the NCut criteria is approximated \cite{ncut_2000} by thresholding each sample's eigenvector corresponding to the second smallest eigenvalue of the Laplacian $\mathbf{L}$, which is defined as:
\begin{eqnarray}
\mathbf{L}= \mathbf{I}-\mathbf{D}^{-1/2}\mathbf{A}\mathbf{D}^{-1/2},
\end{eqnarray}
where matrix $\mathbf{A}$ is an affinity matrix whose entry $A_{ij}$ corresponds to pairwise edge weights $\mathcal{E}$ between samples and $\mathbf{D}$ is a diagonal matrix whose entry $D_{ii}$, defined as $\sum_j A_{ij}$. Thus, instead of optimizing the Eq. 5 directly, we use the Eq. 9 that thresholds $e(I)$ to find the split function $\Psi^*$:
\begin{eqnarray}
\Psi^{*}=\arg\min_{\Psi} \sum_{v\in\{L, R\}} \sum_{I_j\in\mathcal{D}_v} ||e(I_j)-\mu_v||^2
\end{eqnarray}
where $e(I)$ is the eigenvector representation for $I$, corresponding to the second smallest eigenvalue and $\mu_v$ is the mean of $e_j$ for the $v$-th child node.


\paragraph{Higher-order objective: group context consistency.}
 Higher-order potentials \cite{hiercrf1,hiercrf2,hiercrf3} consider relationships among more than $2$ samples. The higher-order objective in Eq. 6 decides the split function $\Psi^*$ that divides samples into two coherent groups $\mathcal{D}_L$ and $\mathcal{D}_R$ where the distances between two group's mean and each sample's value on context features $\mathbf{z}\in\mathcal{Z}$: $||\mathbf{z}(I_j)-\frac{1}{|\mathcal{D}_v|}\sum_{I_l\in\mathcal{D}_v} \mathbf{z}(I_l)||_2, \forall I_l\in\mathcal{D}_v, v\in\{L,R\}$ are minimized. This coincides with the split criterion of regression RFs \cite{hrf_iccv2013,hough,fcode}, which decide a split function having minimum variances of regression labels in left and right child nodes \cite{hrf_iccv2013,hough,fcode}. The same method is used for optimizing the Eq. 6 by regarding the context features as regression labels as in Figure 1d.

\subsection{Testing stage of OAD forests}
\IGNORE{Do we need a separate section here? Can't we make this section a subsection of the previous section? }

Until now, we describe the method for training RFs with the spatio-temporal contexts. In this subsection, we describe the testing stage of RFs for OAD:
\paragraph{Initial action classification.} From each frame $I$, feature vector $\mathbf{x}(I)$ is calculated and inputted to the trained RFs. It propagates through learned split function $\Psi$ of RFs until it reaches to the leaf node at each tree. Note that at the testing stage, only the feature space $\mathcal{X}$ is required since the split function $\Psi(\mathbf{x}_{\gamma}(\cdot),t)$ depends on only $\mathbf{x}\in\mathcal{X}$. The leaf nodes of the forests store class label distributions $P(y\in\mathcal{Y}|\mathbf{x}(I))$ as well as the mean of location features $\mathbf{z}_T$ by using arrived training samples' statistics. By averaging arrived leaf nodes' class label distributions $P(y\in\mathcal{Y}|\mathbf{x}(I))$, we find the inital class label for each frame $I$. Also, location features are averaged over arrived leaf nodes for further processing. 
\paragraph{Action localization.} If the averaged location feature is less than $\beta\in[0,0.5]$ or larger than $1-\beta$, the frame $I$ is likely to locate in start or end frames and we accept action changes in initial class labels. Otherwise, we do not accept any action changes. In this way, the initial  action class labels and the averaged location features are combined to localize action changes. The threshold $\beta$ is decided to minimize the classification error on the training data similar to \cite{Sharaf15}.
\paragraph{Refined action detection.} When there is an action change (\ie start frame), we begin to aggregate initial classification results until when another action change is detected. At the moment when the next action change (\ie end frame) is detected, we stop aggregation and find the most probable action class for the detected interval.

\begin{table}[ht]
\centering
\begin{tabular}{|c|c|c|}
\hline
Method & mean AP & Time per Frame\\
\hline\hline
Moving Pose \cite{movingpose_2013} & 0.890$\pm$0.002 & $-$\\
ELS \cite{Sharaf15} & 0.902$\pm$0.007 & 9.1 ms\\
\hline\hline
RF & 0.820$\pm$0.015 & \\
RF+T & 0.885$\pm$0.012 & 1.1 ms\\
RF+ST & $\mathbf{0.920\pm0.008}$ &\\
\hline
\end{tabular}
\caption{Comparison of performance on MSR Action3D Dataset using the unsegmented setting.}
\end{table}

\begin{table}[ht]
\centering
\begin{tabular}{|c|c|c|}
\hline
Method & avg. F-score & Time per Frame\\
\hline\hline
Bloom \etal. \cite{G3Dtest} & 0.919 & $-$\\ 
Multi-scale \cite{realtime_wacv_2015} & 0.937 & 2.0 ms\\
\hline\hline
RF &  0.887 & \\
RF+T & 0.913 & 1.1 ms\\
RF+ST & $\mathbf{0.948}$ & \\
\hline
\end{tabular}
\caption{Comparison of performance on 'Fighting' Sequence of G3D Dataset based on F1 score at $\Delta=333ms$ using leave-one-out validation.}
\end{table}

\begin{table*}[t]
\centering
\begin{tabular}{|c|c|c|c|c||c|c|c|c|}
\hline
Actions & SVM-SW \cite{online_eccv_2016} & RNN-SW \cite{coocrnn}  & CA-RNN \cite{online_eccv_2016} & JCR-RNN \cite{online_eccv_2016} & RF & RF+T & RF+ST\\
\hline\hline
drinking & 0.146 & 0.441 & $\mathbf{0.584}$ & 0.574 & 0.253 & 0.298 &0.517 \\
eating   & 0.465 & 0.550 & 0.558 & 0.523 & 0.661 & $\mathbf{0.662}$ & 0.645  \\
writing  & 0.645 & 0.859 & 0.749 & 0.822 & 0.761 & $\mathbf{0.858}$ & 0.803 \\
opening cupboard & 0.308 & 0.321 & 0.490 & 0.495 & 0.427 & 0.478 & $\mathbf{0.555}$\\
washing hands & 0.562 & 0.668 & 0.672 & 0.718 & 0.678 & 0.860 & $\mathbf{0.860}$ \\
opening microwave & 0.607 & 0.665 & 0.468 & $\mathbf{0.703}$ & 0.561 & 0.567 & 0.610\\
sweeping & 0.461 & 0.590 & 0.597 & $\mathbf{0.643}$ & 0.224 & 0.273 & 0.437\\
gargling & 0.437 & 0.550 & 0.579 & 0.623 & 0.383& 0.368 & $\mathbf{0.722}$\\
throwing trash & 0.554 & 0.674 & 0.430 & 0.459 & 0.626 & 0.671 & $\mathbf{0.688}$\\
wiping & 0.857 & 0.747 & 0.761 & 0.780 & 0.916 & 0.948 & $\mathbf{0.977}$\\
\hline\hline
Overall F-score & 0.540 & 0.600 & 0.596 & 0.653 & 0.548 & 0.592 & $\mathbf{0.672}$ \\
\hline\hline
Testing Time Ratio & $\mathbf{1.05}$ & 3.14 &  $-$  & 2.60 &  \multicolumn{2}{r}{1.60} &\\
\hline\hline
SL$-$ & 0.316 & 0.366 & 0.378 & 0.418 & 0.320 & 0.356 & $\mathbf{0.445}$ \\
EL$-$ & 0.325 & 0.376 & 0.382 & $\mathbf{0.443}$ & 0.342 & 0.367 & 0.432 \\
\hline
\end{tabular}
\caption{Comparison of performance on Online Action Detection (OAD) Dataset.}
\end{table*}

\section{Experiments}
\label{s:experiments}
We compare our algorithm with eight state-of-the-art algorithms presented in \cite{movingpose_2013,Sharaf15,realtime_wacv_2015,G3Dtest,online_eccv_2016} using three datasets (\ie MSRAction3D \cite{Wanqing_cvprw_2010}, G3D datasets \cite{g3d}, OAD \cite{online_eccv_2016}). Three datasets deal with slightly different OAD problems: OAD without 'no action' frames, OAD with action point annotation and OAD with start/end frame annotation. In Table 1, 2 and 3, we compare our algorithms with state-of-the-art methods in both accuracy and time complexity. 'RF', 'RF+T', 'RF+ST' denote our RF models using no contexts, temporal contexts only, both spatio-temporal contexts, respectively. For our 'RF' model, we use the sliding window approach, where window size is decided by the cross validation as in \cite{movingpose_2013}.

For the RFs, the number of trees is set to $50$ and trees are grown until maximum depth $100$ is reached or minimum sample number $1$ is remained. To maximize the diversity of each tree, we use the bootstrapping method proposed in \cite{bagging} for training each tree of RFs \ie we use different subsets of training data for training each tree. To deal with class imbalance issue \cite{OAD} due to large negative samples (\ie sample that corresponds to 'no-action' class), each subset is sampled from cumulative density function (CDF), whose probability density function (PDF) corresponds to the sample's weight proportional to the inverse of the sample  number in each class. 

In following subsections, we explain three different experimental settings and evaluation protocols using different datasets.

\subsection{MSRAction3D dataset : w/o 'no-action' frames}
\label{es}
 MSRAction3D dataset \cite{Wanqing_cvprw_2010} is originally proposed for the offline classification problem; thus each video is pre-segmented to have only one action class label. In \cite{movingpose_2013}, Zanfir \etal randomly concatenated all $557$ video sequences to one long sequence and use the dataset to mimic the OAD problem; however the dataset is limited in that no negative actions (\ie no action frame) are included. By following the same experimental setting and evaluation protocol as in \cite{Sharaf15,movingpose_2013}, we experiment how well our model deals with the OAD problem without 'no-action' frames. 
 MSRAction3D dataset includes $20$ actions performed by $10$ subjects and the dataset offers depth sequences and $20$ skeleton joints captured by Kinect camera. Since the dataset does not provide RGB frames, we use only depth sequences to constitute the spatial context feature $\mathbf{z}_S\in\mathcal{Z}_S$. 
 
 In Table 1, we compare ours with two methods \cite{movingpose_2013,Sharaf15}. We obtained 'RF' model that is less competitive to both approaches \cite{movingpose_2013,Sharaf15}; however using temporal information, we obtained maximum $6\%$ improvement. Furthermore, by using both spatio-temporal contexts, our framework surpasses state-of-the-art frameworks \cite{movingpose_2013,Sharaf15} by $2\%$. We experiment $100$ random concatenation following \cite{movingpose_2013}; however, the variance of accuracy is small since there is not much difference except for start and end frames that are located between actions.

\subsection{G3D dataset : action point annotation}
G3D dataset \cite{g3d} has {\it action point} annotation. Action point \cite{ap_tr_2012,realtime_wacv_2015} is defined as a single time instance at which the presence of the action is clear, which allows an accurate and fine-grained detection for the actions. By following the same splits and evaluation protocol as in \cite{G3Dtest,realtime_wacv_2015}, we experiment how well our model deals with the OAD problems with action point annotation. In this dataset, RGB frames, depth sequences and $20$ skeleton joint positions captured by the Kinect V1 camera are available. We found that there are many missing RGB frames in this dataset while depth maps are synchronized well with skeletons. We use previous frame's RGB map for the missing frames to constitute the spatial context space $\mathcal{Z}_S$.

In Table 2, we compare ours with two methods \cite{realtime_wacv_2015,G3Dtest} by following their evaluation protocols. Our 'RF' baseline is less competitive than both approaches \cite{realtime_wacv_2015,G3Dtest}; however considering spatio-temporal contexts consistently improve the performance of our model and we obtained the best result.

\begin{figure*}[ht!]
\centering

\subfigure[MSRAction3D dataset]{
   \includegraphics[scale =0.35] {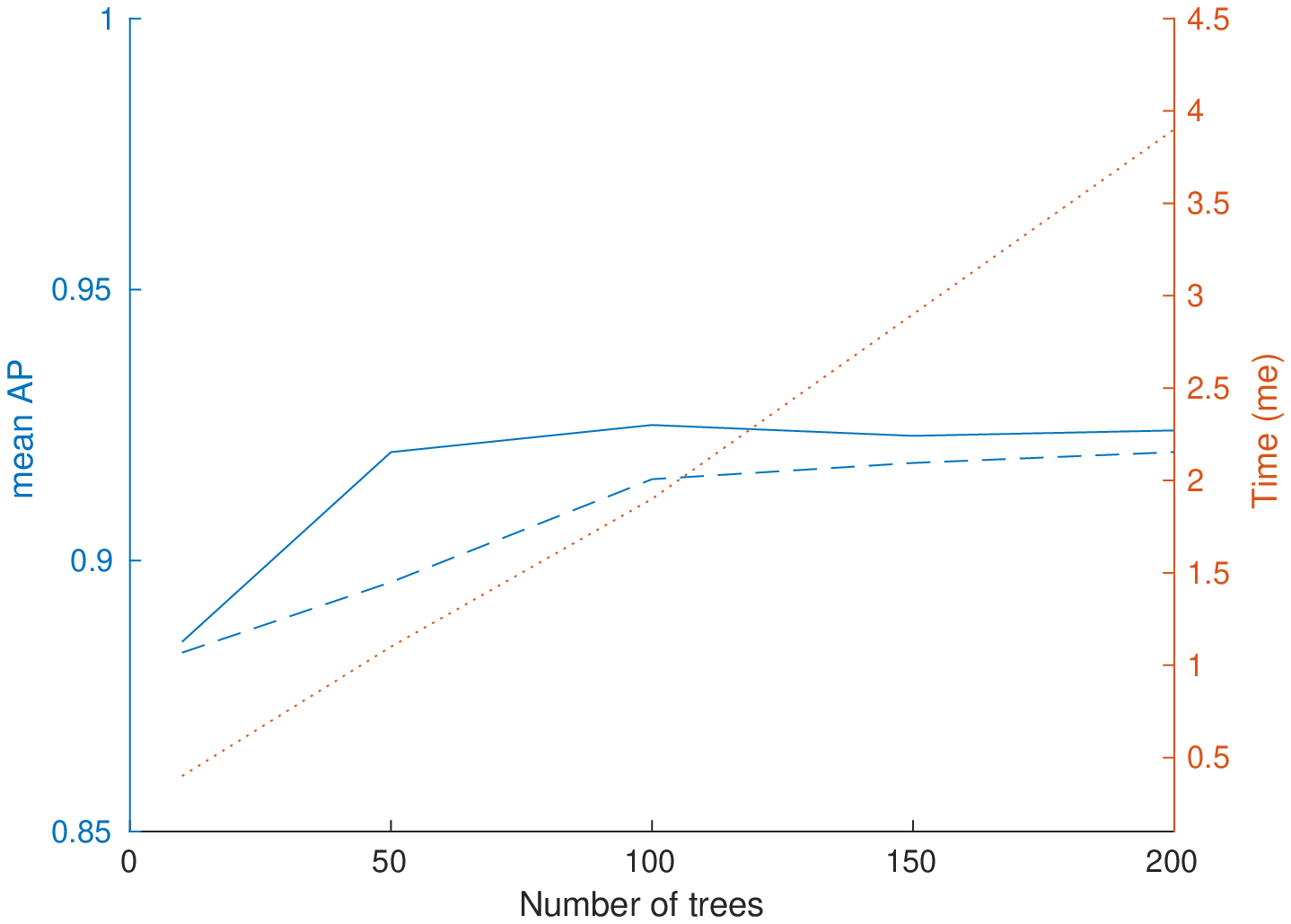}
   \label{fig:subfigg1}
 }
 \subfigure[G3D dataset]{
   \includegraphics[scale =0.35] {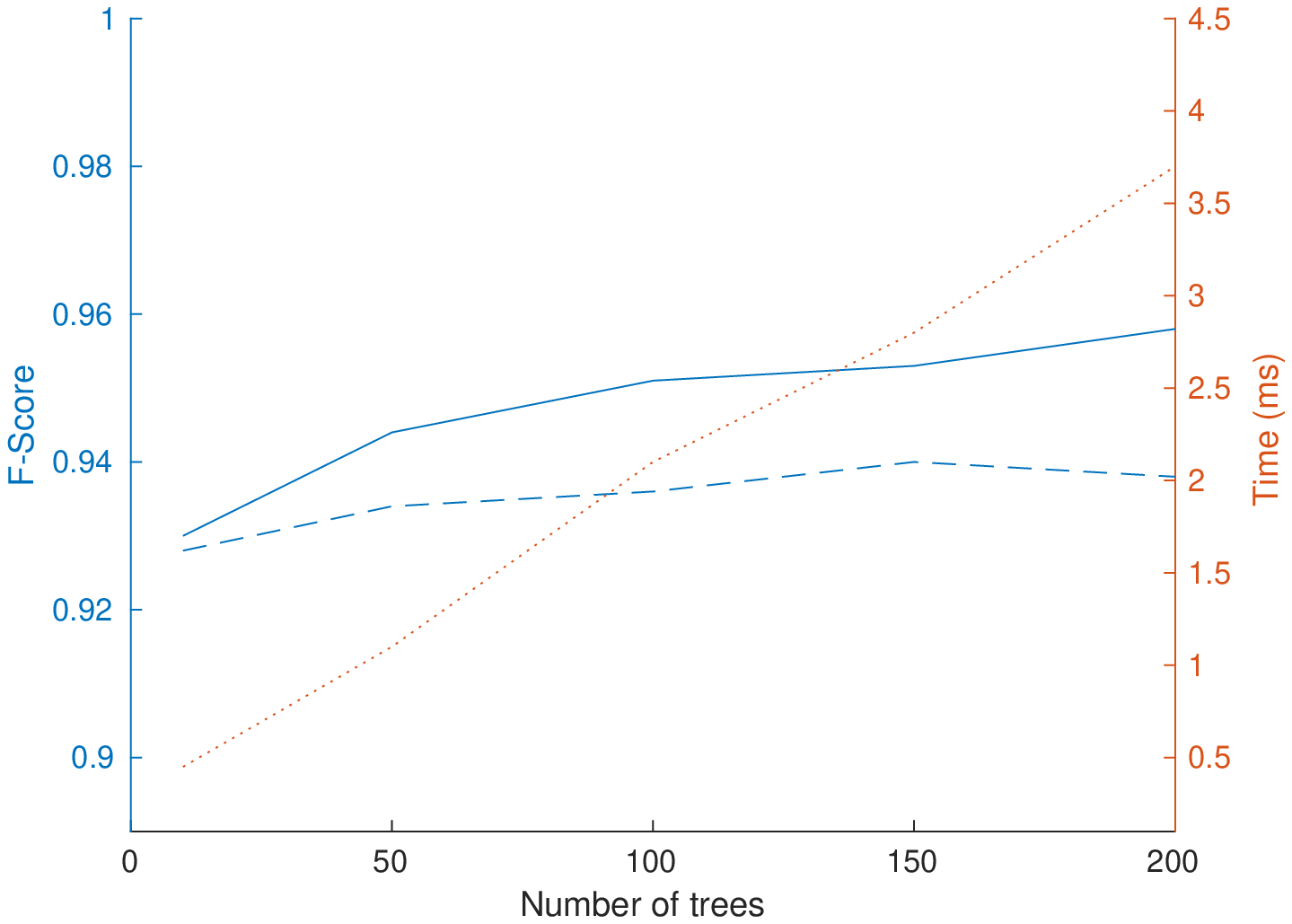}
   \label{fig:subfigg2}
 }
 \subfigure[OAD dataset]{
   \includegraphics[scale =0.35] {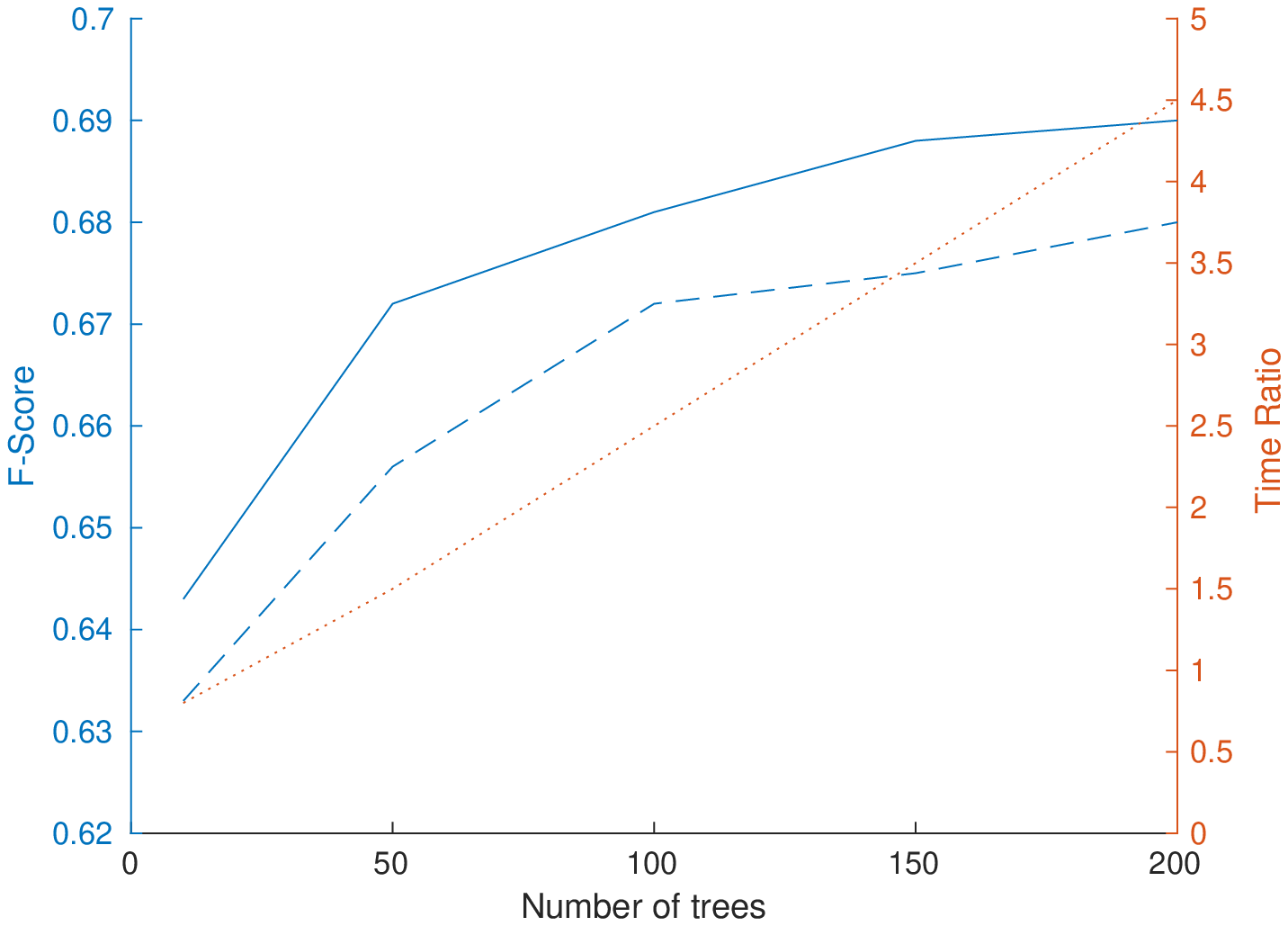}
   \label{fig:subfigg3}
 }
\label{myfigure}
\caption{Graph of Tree numbers vs. Accuracy and Time complexity for each dataset. The blue line and red line plot the change of accuracy and time complexity according to the number of trees, respectively. The dotted blue line denotes the accuracy of RFs without higher-order potentials while the solid blue line denotes that of RFs with higher-order potentials. }
\end{figure*}

\subsection{OAD dataset : start/end frame annotation}
OAD dataset \cite{online_eccv_2016} is recently proposed and in a more challenging setting. It has {\it start/end frame} annotations and has multiple actions in different orders. By following experimental protocols in \cite{online_eccv_2016}, we experiment how well our RFs deal with the OAD problem with start/end frame annotation. The dataset contains $59$ video clips including RGB frames, depth sequences and the tracked $25$ skeleton joint positions, which are captured by the Kinect V2 camera. We use the same splits and evaluation protocol as in \cite{online_eccv_2016}.

In Table 3, we compare ours with four methods in \cite{online_eccv_2016} using their evaluation protocols. They report class-wise F-score as well as overall F-score to measure the accuracy. Also, the accuracy of start and end frame detection is measured by 'SL-' and 'EL-' score that measures the accuracy of start and end frames, respectively. The performance of our 'RF' baseline is similar to that of SVM-SW as both uses sliding window without temporal information. Our 'RF+T' baseline is similar to the RNN-SW of \cite{online_eccv_2016} as both are designed to incorporate temporal information. Our 'RF+ST' baseline achieved the state-of-the-art accuracy which surpasses JCR-RNN model of \cite{online_eccv_2016} in various measures. As discussed in \cite{online_eccv_2016}, the feature space of JCR-RNN is limited to skeletons for the real-time efficiency. Even though RFs are relatively simple compared to RNNs, the incorporation of RGBD modalities by CNNs and temporal information gives us even higher accuracy than RNNs.


\subsection{Time complexity} In Table 1,2, and 3, we measure the time complexity of our method and state-of-the-art methods for MSRAction3D, G3D and OAD datasets, respectively. We measure the testing time using the  6-core 2.84Hz CPU with 32GB of RAM without code optimization. Our method is efficient compared to most state-of-the-arts  \cite{online_eccv_2016,Sharaf15,movingpose_2013,G3Dtest,realtime_wacv_2015} due to two reasons: 1) our feature representation is from \cite{movingpose_2013}, where the low latency is proven. 2) due to the efficient RFs. RFs has the testing time complexity, $O(F\cdot D)$ where $F=50$ is the number of trees and $D=100$ is the maximum tree depth. In the worst case, we have only $F\cdot D$ number of {\it if-statements} for propagating a feature input to the RFs (not a full tree). This is significantly efficient than deep architectures having many filter operations.

In Table 1, the testing time of \cite{movingpose_2013} is unavailable; however, we measure the testing time of \cite{Sharaf15} by their code. In Table 2, the testing time of \cite{G3Dtest} is not available and we measure for \cite{realtime_wacv_2015} using their code. In Table 3, we compare the testing time of methods in \cite{online_eccv_2016} and \cite{coocrnn} indirectly by comparing ours with our own implementation of SVM-SW. We report their testing time ratios rather than the actual timing.

\subsection{Parameter changes}
One of important parameters of RFs for the real-time efficiency is the number of trees. In Figure 2, we plot a graph for the tree numbers vs. accuracy (blue solid) on each dataset. In the same figure, we plot a graph for the tree numbers vs. testing time (red). We also considered the effect of higher-order potentials in the same figure (blue dotted). We notice trade-offs between testing time complexity and the accuracy: More trees in RFs give us improved accuracy, but with more time complexity. Also, we found that the incorporation of higher-order potentials at the training stage consistently gives us a more discriminative RF model.




\section{Conclusion}
\label{s:conclusion}
Skeleton-based OAD is promising due to its efficiency; however it lacks both spatio-temporal contexts. We propose to use CNN feature from RGBDs and temporal location feature as our spatio-temporal contexts.
To incorporate spatio-temporal contexts in the RFs, we utilze the conditional random field-like objective. We effectively optimize it by dividing it and mixing multiple objectives randomly and sequentially in the RFs.\IGNORE{We maintain the efficiency of our testing stage by using contexts to learn more robust RF parameters and do not use any contexts at testing stages.} By experimenting with three datasets, we show that these spatio-temporal contexts significantly improve the  accuracy with low time complexity.

\small
\bibliographystyle{ieee}
\bibliography{egbib}
\end{document}